\documentclass[pdflatex]{sn-jnl}

\jyear{2021}%

\theoremstyle{thmstyleone}%
\theoremstyle{thmstyletwo}%

\theoremstyle{thmstylethree}%

\begin{document}


\title[Article Title]{Exploring a Hybrid Deep Learning Framework to Automatically Discover Topic and Sentiment in COVID-19 Tweets}


\author[1]{\fnm{Khandaker Tayef} \sur{Shahriar}}

\author[1,2]{\fnm{Iqbal} \sur{H. Sarker}}

\affil[1]{\orgdiv{Department of Computer Science and Engineering}, \orgname{Chittagong University of Engineering \& Technology}, \city{Chittagong}, \postcode{4349}, \country{Bangladesh}}

\affil[2]{\orgname{School of Science, Edith Cowan University}, \city{Perth}, \postcode{WA 6027}, \country{Australia}}


\abstract{COVID-19 has created a major public health problem worldwide and other problems such as economic crisis, unemployment, mental distress, etc. The pandemic is deadly in the world and involves many people not only with infection but also with problems, stress, wonder, fear, resentment, and hatred. Twitter is a highly influential social media platform and a significant source of health-related information, news, opinion and public sentiment where information is shared by both citizens and government sources. Therefore an effective analysis of COVID-19 tweets is essential for policymakers to make wise decisions. However, it is challenging to identify interesting and useful content from major streams of text to understand people's feelings about the important topics of the COVID-19 tweets. In this paper, we propose a new \textit{framework} for analyzing topic-based sentiments by extracting key topics with significant labels and classifying positive, negative, or neutral tweets on each topic to quickly find common topics of public opinion and COVID-19-related attitudes. While building our model, we take into account hybridization of BiLSTM and GRU structures for sentiment analysis to achieve our goal. The experimental results show that our topic identification method extracts better topic labels and the sentiment analysis approach using our proposed hybrid deep learning model achieves the highest accuracy compared to traditional models.}

\keywords{COVID-19, Deep Learning, Sentiment Analysis, Topic Identification.}



\maketitle

\section{Introduction}\label{sec1}
Social media plays an important role in a major crisis as people provide feedback and share ideas to others by using these online channels that generate information and new ideas related to disaster response~\cite{o2020social}. Twitter is regarded as one of the most important social media platforms to explain the behavior and predict the outcomes of the pandemic~\cite{jahanbin2020using}. At the end of December 2019, a novel coronavirus outbreak that caused COVID-19 was reported~\cite{malta2020coronavirus}. Due to the rapid spread of the virus, the World Health Organization (WHO) declared a state of emergency~\cite{sohrabi2020world}. The impact of the COVID-19 pandemic on the Twitter platform is huge. Thus, it is very necessary to know the topics related to the COVID-19 tweets and the associating sentiment polarities users are producing in this platform on a regular basis to get an overview of the pandemic situation, human needs, and steps to reduce the harmful impact of the pandemic. The Twitter platform can be considered as a source of useful information to highlight user conversations and better understand public perception towards the COVID-19 pandemic situation. Therefore, by focusing on the context of the COVID-19 pandemic, in this work, we analyze people's tweets on the basis of extracting meaningful topics in an unsupervised manner and predicting sentiments classes in a supervised manner.

Let’s consider a dataset of Twitter that contains user-generated tweets having the sentiment polarities as the target label about COVID-19 related issues. It is very difficult to parse all the tweets and express the internal context manually~\cite{el2019sentiment}. Therefore, finding the internal topics automatically of all tweets will help the policymakers in the relevant departments to implement mandatory measures to reduce the negative impact of the pandemic. However, sentiment analysis is a popular and significant current research paradigm that reflects public perceptions of the event. Sentiment analysis of tweets helps to determine whether a given tweet is neutral, positive, or negative~\cite{mehta2020review}. To achieve our goal, we consider the following research questions:

RQ 1: How to automatically discover the topics associated with the significant labels through analyzing COVID-19 tweets?

RQ 2: How to design an effective framework for predicting the sentiments of tweets with associated topic labels?

To address these questions, we develop a whole new framework by considering the sentiment terms and aspect terms of tweets. However, it is a challenging task to extract meaningful topic labels automatically by machine instead of following the manually annotated topic labeling approach with the diverse human interpretations~\cite{sarker2021data}. Therefore, in this paper, we use LDA~\cite{blei2003latent}, which is an unsupervised probabilistic algorithm to discover topics from tweets. Thus for the topic identification purpose, we do not need any labeled dataset. A collection of topics found in the tweets is generated by LDA. In our earlier paper~\cite{shahriar2021satlabel}, we proposed ``SATLabel" for identifying topic labels comparing the effectiveness with the manual labeling approach only. In this paper, we expand our previous work by changing the LDA-based optimal topic selection process and evaluating the effectiveness with respect to other topic labeling techniques. We also integrate the sentiment detection approach to build a complete framework for topic-based sentiment analysis. Our experimental results show that the label produced by the approach in the proposed framework has the highest soft cosine similarity score with tweets of the same topic compared to other topic labeling methods.

In recent years, deep learning models have become very successful in many fields. Deep learning models automatically extract features from various neural network models and learn from their errors~\cite{zeng2018adversarial}. Deep learning models are extensively used in the field of sentiment analysis by CNN, Recurrent Neural Network (RNN), LSTM, BiLSTM, and Gated Recurrent Unit (GRU)~\cite{zhang2018deep}. GRU is good at capturing order details and long-distance dependency in a sequence from aspect terms and sentiment terms~\cite{zhang2020chinese}. On the other hand, BiLSTM can read the input sequences in the forward and the backward direction to increase the amount of information and improve context successfully~\cite{sharfuddin2018deep}. To improve the performance of sentiment prediction, we propose a hybrid deep learning model based on the combination of the above GRU and BiLSTM features with the use of the Global Average Pooling layer. The primary contributions of this paper can be summarized as follows:

\begin{itemize}
\item We effectively use sentiment terms and aspect terms to identify topic labels in COVID-19 tweets.

\item We propose a hybrid deep learning model combining GRU and BiLSTM features with the use of the Global Average Pooling layer.

\item We have conducted extensive experiments utilizing real-world COVID-19 tweets to show the effectiveness of our proposed framework.
\end{itemize}

The whole paper is formed as follows. Section 2 reviews work on related topic modeling and sentiment analysis. The methodology and architecture of the proposed framework are depicted in section 3. After that, the results of the experiments are analyzed in section 4. Next, we present the discussion and the conclusion section of our work that provide directions for future work.
\section{Related Work}
The suffering caused by COVID-19 is severe~\cite{wang2020phase}. COVID-19 news has been propagated widely on social media surpassing other news. The Tweeter remains on top to broadcast pandemic-related news~\cite{abd2020top}. Tweets related to COVID-19 can be helpful in identifying meaningful topic labels to highlight user conversation and understand ideas about people's needs and interests. Many researchers use the LDA algorithm to extract hidden themes from texts. Hingmire et al.~\cite{hingmire2013document} proposed a paper to create an LDA-based topic model but expert annotation is needed to assign the topic to class labels. Wang et al.~\cite{wang2017hierarchical} presented a paper that minimizes the problem of data sparsity without labeling the main topics particularly. Hourani et al.~\cite{hourani2021arabic} presented a technique of text classification according to their topics that required a labeled dataset. Asmussen et al.~\cite{asmussen2019smart} presented a paper where labeling the topic depends on the researcher's point of view without having any machine-dependent automatic method. Maier et al.~\cite{maier2018applying} introduced the applicability and accessibility of communication researchers using LDA based topic labeling system based on extensive knowledge of the context. Guo et al.~\cite{guo2016big} compared LDA analysis with dictionary-based analysis by implementing a manual topic labeling technique. Ordun et al.~\cite{ordun2020exploratory} labeled each LDA-generated topic with the first three terms having the highest associated probabilities in the corresponding single document of each topic to interpret. Since the Mallet wrapper of the LDA algorithm tends to offer a better division of topics than the inbuilt Gensim version, the LDA Mallet models were implemented in the study~\cite{prabhakaran2018topic} to generate topics where each topic is an integration of keywords and each keyword provides a certain weighted percentage to the topic and the topic labels were inferred from keywords.

Twitter sentiment analysis has become a major source of interest for over a decade~\cite{khurana2018bat}. In history, many works were performed in the field of sentiment analysis using deep learning models~\cite{zhang2018deep, prabha2019survey}. Behera et al.~\cite{behera2021co} presented a hybrid model called Co-LSTM, which highly fits with big social data. Poria et al.~\cite{poria2018multimodal} considered different aspects of analysis in order to investigate multimodal sentiment analysis implementing three deep learning-based structures. Kaur et al.~\cite {kaur2020monitoring} examined emotions and sentiments using TextBlob and IBM Tone analyzer by translating 16,138 tweets into English. Nemes and Kiss~\cite{nemes2021social} scrutinized tweets using TextBlob and RNN. Jiang et al.~\cite{jiang2021toward} proposed a paper based on aspect-level sentiment conversion modules by extracting aspect-specific sentiment words. Imran et al.~\cite{imran2020cross} proposed a multi-layer LSTM based model for predicting both emotions and sentiment polarities. Authors in~\cite{barkur2020sentiment} explored sentiments of Indians in COVID-19 related tweets after the lockdown. For analysis purposes, they collected 24,000 tweets from March 25th to March 28th, 2020 using the keywords: \#IndiaLockdown and \#IndiafightsCorona. The results showed that although there were negative sentiments about the closure, tweets possessing positive sentiments were actually present. Rustam et al.~\cite{rustam2021performance} created the feature set by combining the term frequency-inverse document frequency and bag-of-words for performance comparison of different machine learning techniques by classifying Twitter data as positive, neutral or negative sentiment. They investigated the dataset using the LSTM deep learning model. Naseem et al.~\cite{naseem2021covidsenti} compared conventional machine learning techniques and deep learning methods using the application of various word embeddings such as Glove~\cite{pennington2014glove}, fastText~\cite{bojanowski2017enriching},  and Word2Vec~\cite{mikolov2013efficient} by categorizing COVID-19 related tweets into negative, positive, and neutral classes. Their results suggested that deep learning (DL) methods performed better than traditional ML techniques.

Baired et al.~\cite{stappen2021sentiment} used SenticNet for fine-tuning various features from the dataset. They investigated short microblogging (e.g., Twitter), messaging, and sentiment analysis by combining knowledge-based and statistical methods. A study~\cite{jelodar2020deep} suggested a way to classify sentiments using the deep learning model. It used NLP to model the topic to identify key issues related to COVID-19 expressed on social media. The classification was performed by applying the LSTM Recurrent Neural Networks (LSTM RNN) model. Ahmed et al.~\cite{ahmed2021detecting} used the opinion mining approach to identify sets of users with similar attitudes.  At the beginning of the COVID-19 pandemic, Xue et al.~\cite{xue2020public} performed sentiment analysis on Twitter data expressing that fear of the unknown environment was prevalent on Twitter. Medford et al.~\cite{medford2020infodemic} explored sentiments and investigated topics using LDA by collecting the Twitter data from January 14th to 28th, 2020. Xiang et al.~\cite{xiang2021modern} collected 82,893 tweets for performing sentiment analysis and topic modeling by applying the NRC Lexicon and LDA respectively. Satu et al.~\cite{satu2021tclustvid} presented ``TClustVID" that performs a cluster-based division of tweets and topic modeling. They produced themes of positive, negative, and neutral topics for the cluster investigation and the realization of the COVID-19 pandemic status. Chandrasekaran et al.~\cite{chandrasekaran2020topics} analyzed key topics and corresponding sentiments by identifying the patterns and trends of COVID-19 related tweets before and after the outbreak. They used LDA to extract 26 topics from the COVID-19 related tweets and also clustered them into 10 broad themes to analyze the effect of COVID-19 on various domains.

By analyzing the above works we can reach the conclusion that most of the works do not consider a complete framework for the automatic topic identification and sentiment analysis of COVID-19 related tweets. Moreover, most of the works followed a way of labeling topics manually which is expensive, time-consuming, and dictates difficult human interpretations of the tweets. However, a topic-based analysis of sentiments of the COVID-19 tweets categorizes the predicted multiclass sentiment polarities according to the extracted topics in order to gain an overview of broader public opinion. Therefore, in this paper, we propose a framework for automatically identifying key topics associated with specific topic labels and predicting sentiment polarities to present a topic-based sentiment analysis of the COVID-19 related tweets.
\section{Methodology}
\begin{figure}

\centering
 \includegraphics[width=12cm,height=7cm]{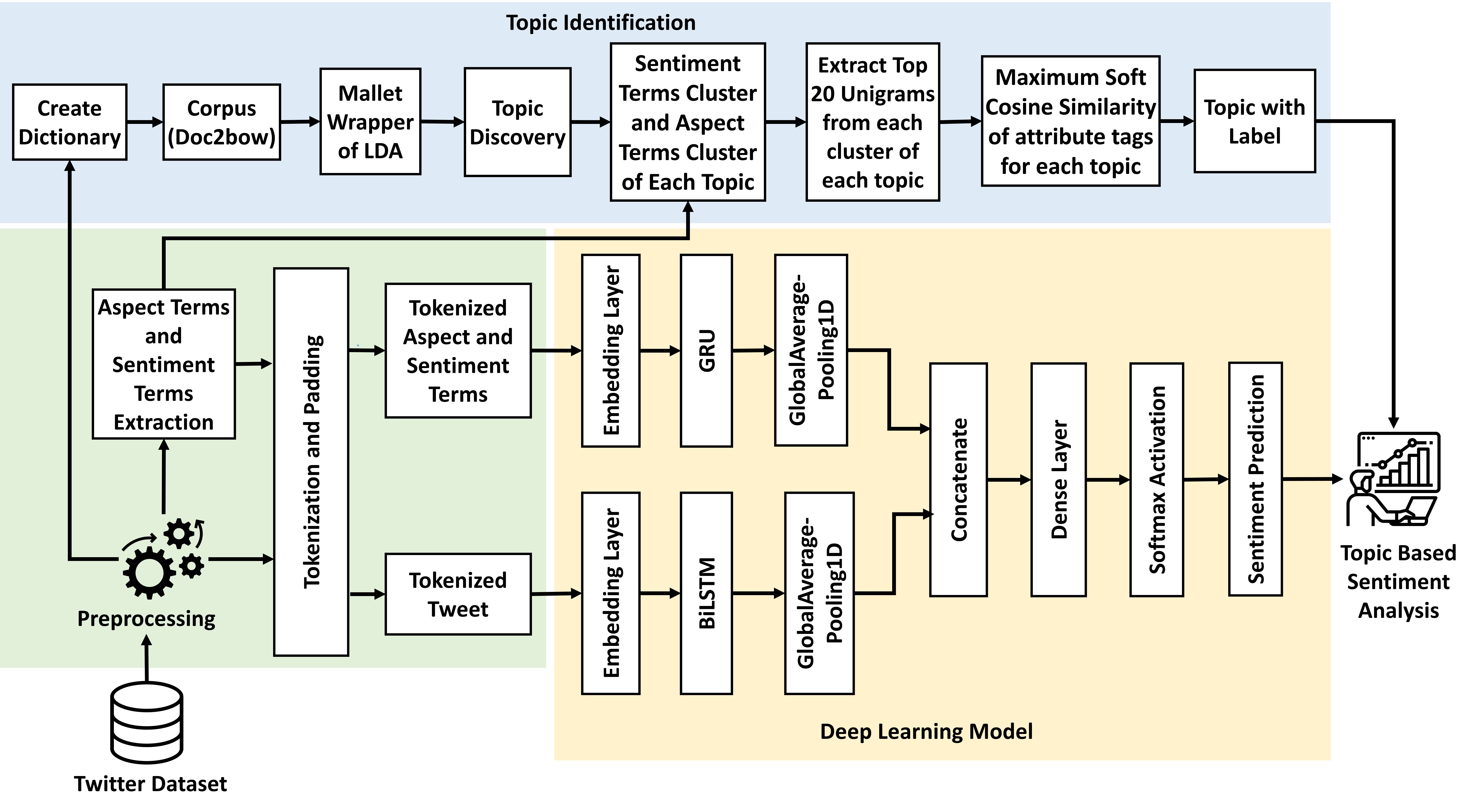}

\caption{Architecture of the proposed framework}
\label{core_framework}

\end{figure}

Topic identification and sentiment analysis (also called opinion mining) play an important role in obtaining information from social media platforms. In this paper, We implement two methods for in-depth analysis of the COVID-19 related tweets on Twitter. The first one is the LDA-based automatic topic modeling and labeling approach, and the second one is multiclass sentiment analysis based on a hybrid deep learning model. Our proposed framework aims to illustrate public sentiments in tweets towards some specific topic labels. In this section, we present our proposed framework for conducting a topic-based sentiment analysis of COVID-19 related tweets as shown in Fig.~\ref{core_framework}. In Fig.~\ref{core_framework}, the ``topic identification" module generates a topic label, and the ``deep learning model" simultaneously predicts the sentiment polarity of each tweet. Therefore, we can perform topic-based sentiment analysis to classify the sentiment polarities of all tweets into corresponding topic labels to provide insight into the COVID-19 pandemic situation. Text pre-processing is one of the most important steps to analyze and extract features from textual data for further processing. After performing normalization of noisy, non-grammatical, and unstructured tweets, we follow a set of processing steps to produce the expected result. By following the topic identification step, our framework separates positive, neutral, or negative sentiments from COVID-19 tweets by implementing a GRU-BiLSTM based hybrid deep learning model. The final summary generated by the proposed framework provides decision support to the policymakers based on the topic-based sentiment analysis which can be presented in any format, either text or charts.
\subsection{Preparing Dataset}
TextBlob, VADER and other sentiment analysis tools are often used to detect sentiments from COVID-19 related tweets~\cite{rustam2021performance,ahmed2021detecting,nemes2021social,jiang2021toward}. But in this work, we only consider the publicly available manually tagged dataset with sentiment polarities of the COVID-19 related tweets from the Kaggle repository\footnote{\url{https://www.kaggle.com/datatattle/covid-19-nlp-text-classification}}. There are two CSV files in the dataset. One is Corona\_NLP\_train.CSV and another is Corona\_NLP\_test.CSV. There are  41,157 tweets in Corona\_NLP\_train.CSV file and 3,798 tweets in Corona\_NLP\_test.CSV file. There are five categories of sentiments as Positive, Extremely Positive, Neutral, Negative, and Extremely Negative which are transformed into three classes namely Positive, Neutral and Negative as a target label to achieve better accuracy. There are 7,713 Neutral, 15,398 Negative, and 18,046 Positive tweets available in the dataset which are marked with 0, 1, and 2 respectively. We prepare the Twitter dataset to get the standard form by using several text processing functions.
\begin{enumerate}
\item[{\it 1)}]{\it Conversion of words into lowercase:}
We transform all words in the tweets into lowercase.
\item[{\it 2)}]{\it Deletion of URLs and links:}
We replace all the URLs and hyperlinks in the tweets with empty strings.
\item[{\it 3)}]{\it Elimination of mentions:}
Normally people use @ in tweets to mention users. In our work, we eliminate the mentions.
\item[{\it 4)}]{\it Removal of hashtags:}
We identify hashtags and remove them with empty strings because hashtag words do not carry useful meaning in sentiment analysis~\cite{hashtag}.
\item[{\it 5)}]{\it Handling contractions:}
Generally, people use common English contractions in social media. We apply the regular expression to handle common contractions. For example, we replace won't with would not.
\item[{\it 6)}]{\it Elimination of punctuations:}
There is no importance of punctuations in tweets. We withdraw punctuation symbols such as \&, \_, -, etc.
\item[{\it 7)}]{\it Managing words less than two characters:}
We remove the words whose length is less than two characters because the word such as `I' does not really carry much influence in determining sentiment.
\item[{\it 8)}]{\it Stop words elimination:}
Stop words in the English language such as pronouns, articles, prepositions, etc. have no emotional impact on the sentiment classification process. We remove the stop words.
\item[{\it 9)}]{\it Dealing with Unicode and Non-English words:}
To develop a clean and noise-free dataset, we remove the tweets that are not in English, and Unicode like ``\textbackslash u018e".
\item[{\it 10)}]{\it Tokenization:}
We vectorize the corpus by transforming each text into a sequence of integers based on word count.
\end{enumerate}
\subsection{Extracting Sentiment Terms and Aspect Terms}
Sentiment terms capture the tone or perception of a sentence. Usually, adjectives and verbs are regarded as sentiment terms in a sentence that reflects the expressed opinion of the text. Noun phrases and nouns are regarded as aspect terms. Aspect terms are often considered as features that describe the event, product, or entity~\cite{wang2017coupled}. We follow the precise parts of speech tagging which is an effective way to extract sentiment terms and aspect terms from texts. Examples of sentiment terms and aspect terms are shown in Table~\ref{table1}.
\begin{table}
\caption{\textbf{Examples of Sentiment Terms and Aspect Terms}}
\centering
\label{table1}
\begin{tabular}{|c|c|c|}
\hline
\textbf{Sample Tweet} &
  \textbf{\begin{tabular}[c]{@{}c@{}}Sentiment\\ Terms\end{tabular}} &
  \textbf{\begin{tabular}[c]{@{}c@{}}Aspect\\     Terms\end{tabular}} \\ \hline
Please read the thread. &
  read &
  thread \\ \hline
\begin{tabular}[c]{@{}c@{}}To enjoy and relax for your dinner\\ it is a great place.\end{tabular} &
  \begin{tabular}[c]{@{}c@{}}enjoy, relax,\\ great\end{tabular} &
  dinner, place \\ \hline
\begin{tabular}[c]{@{}c@{}}Links with info on communicating with\\ children regarding COVID-19.\end{tabular} &
  \begin{tabular}[c]{@{}c@{}}communicate,\\ covid\end{tabular} &
  \begin{tabular}[c]{@{}c@{}}links, info,\\ children\end{tabular} \\ \hline
The retail store owners right now &
  retail, right &
  owners, store \\ \hline
\end{tabular}
\end{table}
\subsection{Feature Extraction}
To implement the machine learning and deep learning models, feature extraction is essential from the tweets~\cite{sarker2021machine, sarker2021deep}. Two methods of extracting features, a bag of words (BoW) and Word2vec are utilized in the proposed framework.

The boW is an easy way to extract features and it is used to retrieve information and perform NLP tasks~\cite{rustam2020classification}. BoW generates features to train the ML models on the basis of the presence of the individual words in the text and is widely used in text classification tasks. BoW produces vocabulary for all unique words and their frequency of occurrences in the documents to train models. In our work, we implement Doc2bow to extract numerical features from tweets to train the LDA model, which is a simple technique in Gensim~\cite{habibabadi2019topic} that converts documents into a representation of the bag of words as shown in the topic identification section of Fig.~\ref{core_framework}. In this paper, we use word2vec (skip-gram version) as embedding layers to extract numerical features from tokenized texts and develop our framework for sentiment analysis purposes~\cite{mikolov2013efficient,jang2019word2vec}.
\begin{algorithm}[H]
\DontPrintSemicolon
  
  \KwIn{T: number of Tweets in dataset}
  \KwOut{Topic Label ($T_{Label}$).}
\For{\textbf{each} t $\in$ \{1,2,...,T\} }{
  $ T_p \leftarrow Preprocess(t);$
}
initialize list $Corpus$\\
  \For{\textbf{each} $t_p$ $\in$ \{1,2,...,$T_p$\} }{
  \tcp{Corpus Development}
  append dictionary(Doc2bow($t_p$)) to $Corpus$;
  
  \tcp{Sentiment Terms and Aspect Terms Extraction}
  $S_{T_p} \leftarrow Sentiment\_Terms(t_p);$
  
  $A_{T_p} \leftarrow Aspect\_Terms(t_p);$
 }
  \tcp{Topic Discovery}
  $K \sim Mallet(LDA(Corpus));$
  
  \For{\textbf{each} k $\in$ \{1,2,...,K\} }{
  
  \For{\textbf{each} $t_p$ $\in$ \{1,2,...,$T_p$\} }{
  $k_{dominant,t_p} \sim dominant\_topic(t_p,k);$ 
  
  \tcp{Create Clusters from Topic} $C_S \sim Cluster( S_{T_p}\rightarrow k_{dominant,t_p});$
  
  $C_A \sim Cluster( A_{T_p}\rightarrow k_{dominant,t_p});$
  }}
  \For{\textbf{each} k $\in$ \{1,2,...,K\} }{
  \tcp{Extract top 20 unigrams from sentiment terms cluster of each topic}
  $U_S\sim max\_count(Top\_Unigrams(C_S\rightarrow k));$\\
  \tcp{Extract top 20 unigrams from aspect terms cluster of each topic}
  $U_A\sim max\_count(Top\_Unigrams(C_A\rightarrow k));$
  
  \tcp{Create Attribute Tags}
  
  $A_{tag}\leftarrow U_S + U_A$
  
  $T_{Label}\leftarrow max\_soft\_cosine\_similarity(A_{tag}, k)$
  }
\caption{Automatic Topic Labeling}
\end{algorithm}
\subsection{Topic Identification}
LDA is a well-performed topic modeling algorithm for finding hidden themes available in the corpus on an unlabeled dataset~\cite{sarker2021machine}. But the challenge is how to assign important labels on topics generated by the LDA. The working principle for generating automatic topic labels from the Twitter dataset is presented in Algorithm 1. The steps for the identification of topics and the generation of significant topic labels automatically as output in the proposed framework are discussed below:
\subsubsection{Creating Dictionary and Corpus}
The systematic way of generating a number of language lexicons is supported by a dictionary and the corpus usually illustrates an arbitrary sample of that language. The corpus of a document is made up of words or phrases. In the Natural Language Processing (NLP) paradigm, the language corpus plays an important role in creating a knowledge-based system and text mining. In the proposed framework, we construct a dictionary from the preprocessed text and then create a corpus. Documents in the dictionary are converted to the Bag of Words (BoW) format using Doc2bow embedding~\cite{habibabadi2019topic} for the corpus development. Corpus comprises the words id and its frequency in all documents.  Each word is considered as a normalized and tokenized string.
\subsubsection{Topic Discovery}
The BoW corpus is transferred to the LDA mallet wrapper. The presence of a set of topics on the corpus is discovered by the LDA. LDA Mallet wrapper works fast and provides a better classification of topics using the Gibbs Sampling~\cite{boussaadi2020researchers} method. LDA produces the most prominent words in the themes. So by using the weightage of keywords one can infer prevalent topics in texts. In order to overcome the difficult manual labeling method, our framework produces automated topic labels using sentiment terms and aspect terms without human interpretation. Based on the result of the topic coherence score, we select an LDA model that gets a total of 14 topics itself. Then we get the main topic as dominant topic in each tweet from LDA model to identify the distribution of topics across all tweets in the dataset.
\subsubsection{Labeling Topic Using Top Unigrams}
We generate sentiment terms cluster and aspect terms cluster by experimenting with tweets corresponding to each LDA-generated topic. Therefore, we find 14 sentiment terms cluster and 14 aspect terms cluster from LDA-generated topics. Unigram is a single n-gram word sequence. The use of unigrams can be seen in NLP, cryptography, and mathematical analysis. However, the soft cosine similarity takes into account the similarity of the features in the vector space model~\cite{sidorov2014soft}. For each topic, we select top 20 unigrams as threshold value because of high frequencies of these unigrams and extract the top 20 unigrams from the sentiment terms cluster and the top 20 unigrams from the aspect terms cluster. Then we create all the possible combinations of the top 20 unigrams of sentiment terms with the top 20 unigrams of aspect terms for each topic respectively to generate an appropriate attribute tag. We choose an attribute tag that has the highest soft cosine similarity value in relation to the tweets of that topic to assign with a significant topic label. We use sentiment term and aspect term blended attribute tag to label each topic because that feature tag describes the topic of the tweets. To classify a tweet into categories with a specific topic label from test data, we detect the topic number that contains a significant percentage impact on that tweet.
\subsection{Building a Hybrid Deep Learning Model for Sentiment Analysis}
In our proposed framework we implement a hybrid deep learning model for multiclass sentiment classification using two powerful feature extractors. Gated Recurrent Unit (GRU) and Bidirectional Long Short Term Memory (BiLSTM) are used for feature extraction in two branches of the hybrid model in the proposed framework as shown in Fig.~\ref{core_framework}. Both of the extractors are followed by an individual Global Average Pooling layer. In embedding layers, we apply the word embeddings generated by Word2vec (skip-gram version) model to measure the semantic relationship between each word pair to represent. The GRU branch takes tokenized sentiment terms and aspect terms as input, and the BiLSTM branch takes the tokenized tweet as input. The concatenation layer merges the features coming from two branches followed by the fully connected dense layer with softmax activation function for multiclass sentiment classification i.e. positive, neutral, or negative as shown in Fig.~\ref{core_framework}. Below we describe the main models used to build the GRU-BiLSTM based hybrid model in the proposed framework:

Long Short Term Memory (LSTM) is a kind of RNN architecture that handles the vanishing gradient problem using exclusive units~\cite{hochreiter1997long}.  Data is stored for a long time in the memory cell of the LSTM unit and three gates control the access to information inside and outside the cell. For instance, which information from the previous state cell will be memorized and which information will be deleted is determined by ``Forget Gate", while which information should enter the cell state is determined by ``Input Gate", and the output is controlled by the ``Output Gate". Bidirectional LSTM, more commonly known as BiLSTM, is a standard LSTM extension that enhances the model performance in the order of sequence~\cite{siami2019performance}. It is a type of sequence processing model combining two LSTMs: one receives the input approaching in the forward direction and the other receives it in the backward direction. In natural language processing activities, Bidirectional LSTM is especially considered a popular option to carry on.

Gated Recurrent Unit (GRU) is another kind of recurrent network where information flow is managed and controlled between cells by implementing gating techniques in the neural network~\cite{sarker2021deep}. GRU contains an update gate and a reset gate without having an output gate. The GRU structure makes it possible to take on the dependence of large sequences of data adaptively, without losing any data from earlier segments of the sequence. Moreover, GRU is a relatively simple type and much faster to calculate which often provides comparable functionality. The performance of the GRUs is better on certain smaller and less frequent datasets~\cite{sarker2021deep}.

Global Average Pooling (GAP) does not slide in the structure of a small window but is measured over the entire output feature map in the previous layer. GAP typically regularizes the whole network structure, and each output channel corresponds to the features of each class, making the relationship between the output and the feature class more intuitive. Moreover, the GAP layer contains no data parameter. Therefore, using GAP helps to improve network performance and increase cognitive accuracy~\cite{li2021multi}. In addition, the GAP layer avoids overfitting because it does not need parameter optimization. It serves as a flatten layer for transforming a multi-dimensional feature space into a one-dimensional feature map. In addition, it takes less time for computation.
\begin{table}[]

\centering
\caption{\textbf{The model hyper-parameters}}
\label{hyperparameter}
\begin{tabular}{|c|c|c|c|}
\hline
\textbf{Hyper-parameter}                 & \textbf{\begin{tabular}[c]{@{}c@{}}Initial\\ Value\end{tabular}} & \textbf{\begin{tabular}[c]{@{}c@{}}Hyper-parameter\\ Space\end{tabular}} & \textbf{\begin{tabular}[c]{@{}c@{}}Optimal  \\ Value\end{tabular}} \\ \hline
Word embedding size   for word2vec       & 100                                                              & 50, 100, 150                                                             & 100                                                                \\ \hline
The number of   hidden neurons in BiLSTM & 128                                                              & 64, 128, 256, 512                                                        & 256                                                                \\ \hline
The number of   hidden neurons in GRU    & 128                                                              & 64, 128, 256                                                             & 128                                                                \\ \hline
Batch Size                               & 32                                                               & 16, 32, 64                                                               & 32                                                                 \\ \hline
\end{tabular}

\end{table}

\begin{figure}

\centering
\includegraphics[width=12cm,height=12cm]{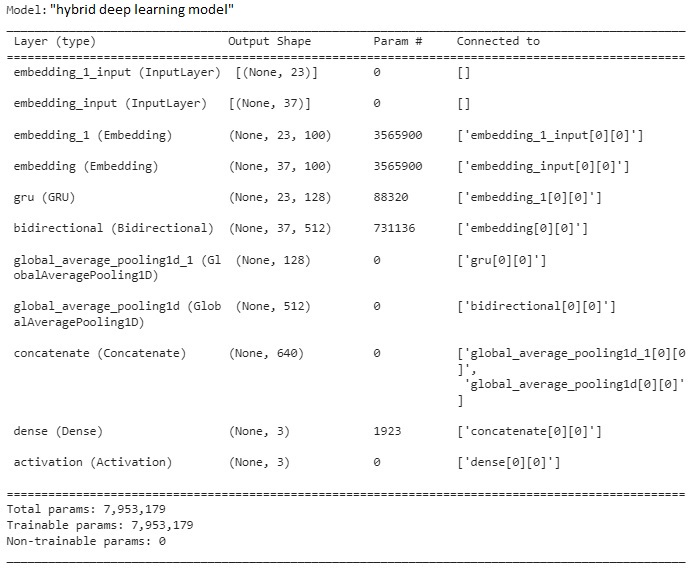}
\caption{Model summary for hybrid deep learning model.}
\label{model_summary}

\end{figure}

Our hybrid deep learning-based model in the proposed framework effectively utilizes the advantages of BiLSTM, GRU, and GAP models for multiclass sentiment classification. To predict the sentiment polarities, we apply a fully connected dense layer to extract the features from the feature map of the concatenation layer as shown in Fig.~\ref{core_framework}. The Softmax classifier acts as its input and takes the output to the final step. Table~\ref{hyperparameter} shows the hyper-parameters used in the proposed hybrid model. As summarized in Table~\ref{hyperparameter}, the hyper-parameters characteristics include word embedding size of word2vec, number of hidden neurons in BiLSTM, GRU, and batch size. Due to the different lengths of tweets in the dataset, we take the maximum length of the tweets when we input tweets to the model. RMSProp optimizer is used because RMSProp incorporates the best properties to adjust learning rate adaptively and optimize model parameters~\cite{kamila2019resolution}. The pseudo-code of the proposed hybrid deep learning model in the framework is provided in Algorithm 2. The model summary for the hybrid deep learning model is as shown in Fig.~\ref{model_summary}. The shape of the output of the hybrid deep learning model is 3 as seen in the last block of Fig.~\ref{model_summary}.
\begin{algorithm}[H]
\DontPrintSemicolon
  
  \KwIn{COVID-19 Training Set T1, Testing Set T2}
  \KwOut{Predicted sentiment classes which are $s_{pos}, s_{neu}, s_{neg}$}
Preprocess Tweets\\
Extract tokenized sentiment terms and aspect terms $tok_{sa}$\\
Extract tokenized tweet $tok_{tweets}$ \\
 Initialize hyperparameters for word2vec model\\
 Train word2vec embedding model using T1\\
 Initialize hyperparameters for GRU and BiLSTM\\
 Set $\leftarrow$ embedding layer, GRU layer, GlobalAveragePooling Layer in one branch for $tok_{sa}$\\
 Set$\leftarrow$ embedding layer, BiLSTM layer, GlobalAveragePooling Layer in another branch for $tok_{tweets}$\\
 Concatenate the features coming from two branches\\
 Set$\leftarrow$ dense layer with softmax activation\\
 Train the hybrid model and compute the initial weights\\
  \For{\textbf{epoch} 1 to max }{
  Select a mini batch from T1\\
  Forward propagation and compute the loss via learning rate\\
  Backpropagation and update weights with RMSProp optimization\\
 }
  $[s_{pos}, s_{neu}, s_{neg}]$ = hybrid model(T1, T2)
\caption{Pseudo code of the proposed hybrid model}
\end{algorithm}
\subsection{Topic Based Sentiment Analysis}
After predicting the sentiments of the tweets, the final step of our proposed framework is to classify the sentiment polarities for each of the 14 topics. We visualize and count the number of positive, negative, and neutral tweets to provide a summary to support decision-making for the policymakers as shown in Fig.~\ref{topic_based_sentiment}. Our proposed framework effectively conducts topic-based sentiment analysis for providing a comprehensive overview of the broader public opinion. In the experimental section, we demonstrate the performance and effectiveness of our framework with respect to other benchmark models.
\section{Experimental Results}
In this section, we present the outcomes and findings of our experiments. First, we search trending topics with specific labels. Then we perform sentiment analysis of the tweets using a hybrid deep learning model. We also make the comparison of the results of the proposed framework with the corresponding benchmark models.
\subsection{LDA Based Topic Identification}
\subsubsection{Finding The Optimal Number of Topics for LDA}
\begin{figure}
\centering
\includegraphics[width=21.5pc,height=6cm]{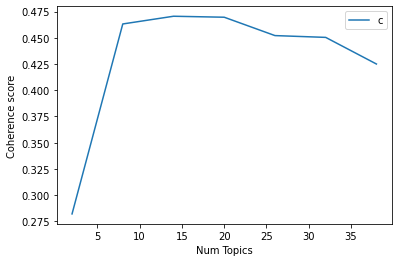}
\caption{Coherence score for the number of topics.}
\label{coherence_score}
\end{figure}
We build a function to extract a number of LDA models with multiple coherence values for the number of topics in order to obtain the optimal number of topics. LDA produces common co-occurred words and builds them into separate themes. We choose the optimal number of topics on the basis of the gensim coherence model~\cite{xue2020public}. We select the LDA model that returns the number of 14 topics for this dataset because it generates the highest coherence value as presented in Fig.~\ref{coherence_score}. Fig.~\ref{coherence_score} shows the graph of coherence scores for the different number of topics returned by the multiple LDA models.
\subsubsection{Selection of Top Unigram Features from Clusters}
\begin{figure}
\begin{center}
\includegraphics[width=\textwidth, height=5.5cm]{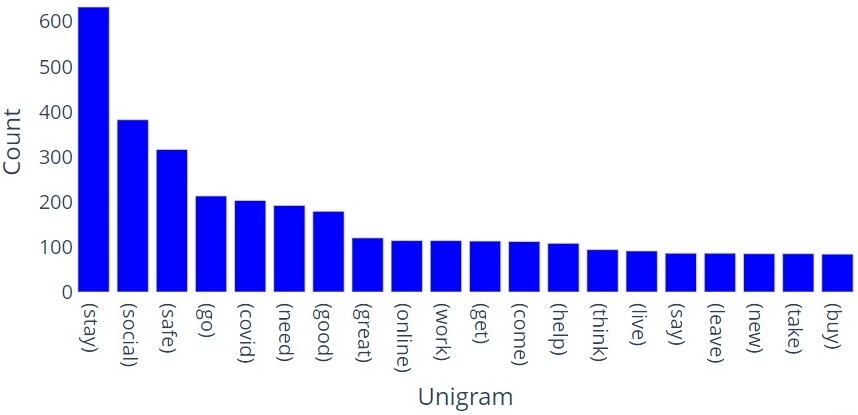}
\caption{Top 20 unigrams from sentiment terms cluster of topic no. 3}
\label{3_sentiment}
\end{center}
\end{figure}
\begin{figure}
\begin{center}
\includegraphics[width=\textwidth, height=5.5cm]{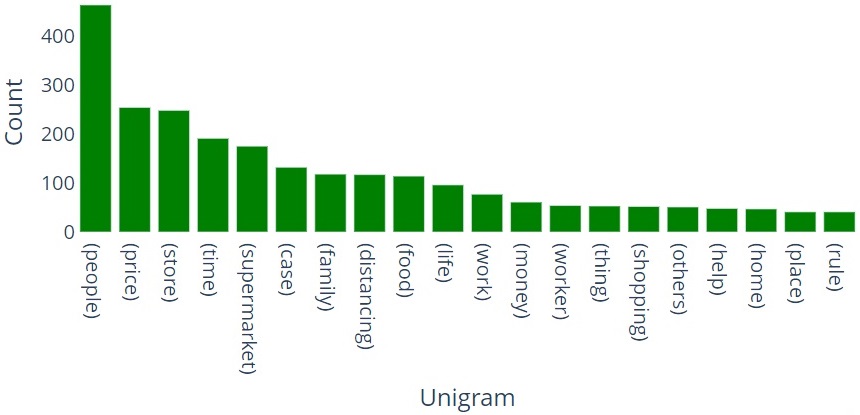}
\caption{Top 20 unigrams from aspect terms cluster of topic no. 3}
\label{3_aspect}
\end{center}
\end{figure}
We generate sentiment terms cluster and aspect terms cluster for each topic. We find at the top count 20 unigrams in each cluster. Fig.~\ref{3_sentiment} and Fig.~\ref{3_aspect} show the top 20 unigrams from sentiment terms cluster and aspect terms cluster of the topic no. 3 respectively. Then we get the topic label depending on the highest soft cosine similarity value of the attribute tag in relation to the tweets for that particular topic, which is generated by the combination of top unigrams of sentiment terms and aspect terms.
\subsubsection{Qualitative Evaluation of Topic Labels}
In table~\ref{topic_label}, we present part of a set of tweets assigned by the proposed framework generated topic labels. Table~\ref{topic_label} presents that the detected topic labels are well-aligned and closely coherent with the descriptions of the tweets. We can categorize tweets and extract useful information related to a topic, by simply separating the tweets using the key label generated by the proposed method for that topic.
\begin{table}[]
\centering
\caption{{\bf Example of Topics Detected on Tweets.}}
\label{topic_label}
\begin{tabular}{|c|c|c|}
\hline
\textbf{Sample Tweet}                                                                                                                                                                                                                                                                                                                          & \textbf{Topic No} & \textbf{Topic Label} \\ \hline
\begin{tabular}[c]{@{}c@{}}create contact list with phone numbers of neighbours schools\\ employer   chemist GP set up online shopping accounts\\ if poss adequate supplies of   regular meds but not over order\end{tabular}                                                                                                                  & 5                 & online slot                   \\ \hline
\begin{tabular}[c]{@{}c@{}}Coronavirus Australia: Woolworths to give elderly,\\ disabled dedicated   shopping hours amid COVID-19 outbreak\end{tabular}                                                                                                                                                                                        & 1                 & thank store                   \\ \hline
\begin{tabular}[c]{@{}c@{}}My food stock is not the only one which is empty...\\ PLEASE, don't panic, THERE WILL BE ENOUGH\\ FOOD FOR EVERYONE if you do   not take more\\ than you need. Stay calm, stay safe.\end{tabular}                                                                                                                   & 2                 & hoard food                    \\ \hline
\begin{tabular}[c]{@{}c@{}}Dear Coronavirus,\\ I've been   following social distancing rules and staying home\\ to prevent the spread of   you.  However, now I've spent an\\ alarming amount of money shopping online.    Where can I\\ submit my expenses to for reimbursement? Let me know.\\  \#coronapocolypse \#coronavirus\end{tabular} & 5                 & safe distancing               \\ \hline
\begin{tabular}[c]{@{}c@{}}Cashier at grocery store was sharing his insights\\ on \#Covid\_19 To   prove his credibility he commented\\ "I'm in Civics class so I know what   I'm talking about".\end{tabular}                                                                                                                                 & 10                & learn behavior                \\ \hline
\begin{tabular}[c]{@{}c@{}}Due to COVID-19 our retail store and classroom\\ in Atlanta will not be   open for walk-in business\\ or classes for the next two weeks, beginning   Monday,\\ March 16.  We will continue to   process online and\\ phone orders as normal! Thank you for your understanding!\end{tabular}                         & 6                 & small event                   \\ \hline
\begin{tabular}[c]{@{}c@{}}Airline pilots offering to stock supermarket shelves in\\ \#NZ lockdown   \#COVID-19\end{tabular}                                                                                                                                                                                                                   & 12                & covid supermarket             \\ \hline
\begin{tabular}[c]{@{}c@{}}@TartiiCat Well new/used Rift S are going for \$700.00\\ on Amazon rn   although the normal market price is\\ usually \$400.00 . Prices are really crazy   right now for\\ vr headsets since HL Alex was announced and it's only\\ been   worse with COVID-19. Up to you whethe\end{tabular}                          & 11                & crude market                  \\ \hline
\begin{tabular}[c]{@{}c@{}}Response to complaint not provided citing COVID-19\\ related delays.   Yet prompt in rejecting policy before\\ consumer TAT is over. Way to go ?\end{tabular}                                                                                                                                                       & 7                 & learn scam                    \\ \hline
\end{tabular}

\end{table}
\subsubsection{Effectiveness Analysis}
\begin{figure}

\centering
\includegraphics[width=12cm,height=6cm]{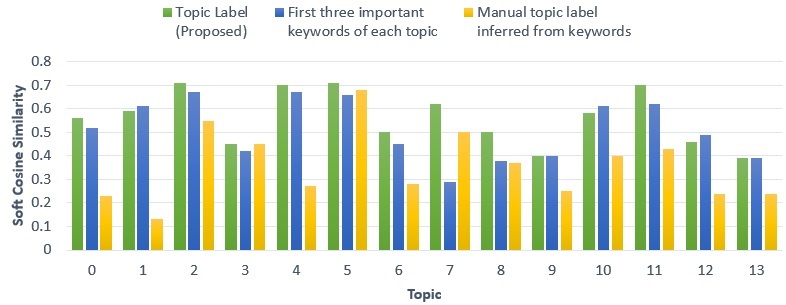}
\caption{Comparison of proposed topic label approach with other approaches}
\label{scs_value}

\end{figure}

In this experiment section, we compare the Soft Cosine Similarity (SCS) values of topic labels obtained by our proposed method with other two approaches as shown in Fig.~\ref{scs_value}. One is the approach followed by the first three important keywords of each topic~\cite{ordun2020exploratory} for topic labeling and another is the manual topic label inferred from keywords for LDA-generated 14 topics~\cite{prabhakaran2018topic} as LDA generates keyword importance percentages for each topic. SCS is used to find semantic text similarities between two documents. The high value of SCS provides a high degree of similarity and the low SCS value provides less similarity to unrelated documents.

From Fig.~\ref{scs_value}, we find that SCS values generated by the proposed method for all topics are higher than other approaches except topics no. 1, 10, and 12. In the case of these three topics, the approach followed by Ordun et al.~\cite{ordun2020exploratory} finds slightly higher SCS values. For topics no. 9 and 13 we find the same SCS values for our proposed method and method followed by Ordun et al.~\cite{ordun2020exploratory}. In case of topic no. 3 we find the same SCS value of the method followed by Prabhakaran~\cite{prabhakaran2018topic} with our proposed method. From Fig.~\ref{scs_value}, we can see that the topic labels produced by the proposed method in the framework illustrate higher SCS values for a maximum number of topics compared to other labeling methods. Therefore, our proposed framework works better and traces better topic labels from the dataset in an unsupervised manner to reduce the complex workload of human labeling.
\subsection{Sentiment Analysis Using Hybrid Deep Learning Model}
\subsubsection{Effect of Model Iterations}
\begin{figure}
  \includegraphics[width=.8\linewidth]{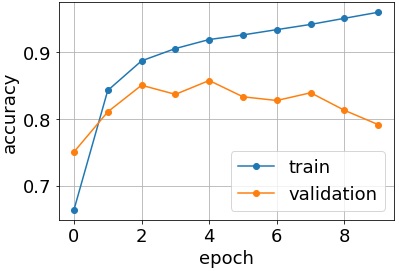} 
  \caption{Model accuracy graph.}
  \label{model_accuracy}
\end{figure}

\begin{figure}
  \includegraphics[width=.8\linewidth]{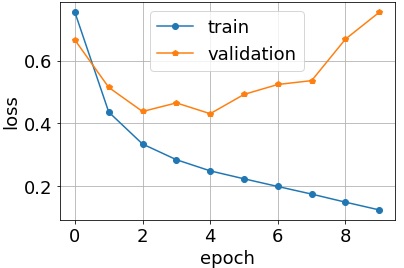}  
  \caption{Model loss graph.}
  \label{model_loss}
\end{figure}

Multiple iterations of model training affect model performance. The validation accuracy of the model first rises and then falls with increasing model iterations. It can be seen in Fig.~\ref{model_accuracy} and Fig.~\ref{model_loss}, when the number of iterations of the model increases, the model gradually overfits, and performance decreases. Fig.~\ref{model_accuracy} and Fig.~\ref{model_loss} present the model accuracy and loss graph of the proposed hybrid deep learning model in the framework respectively. From Fig.~\ref{model_accuracy} and Fig.~\ref{model_loss}, we find the best model at epoch no. 4.

\subsubsection{Performance Comparison of Different Models}
\begin{figure}
\includegraphics[width=\textwidth,height=5cm]{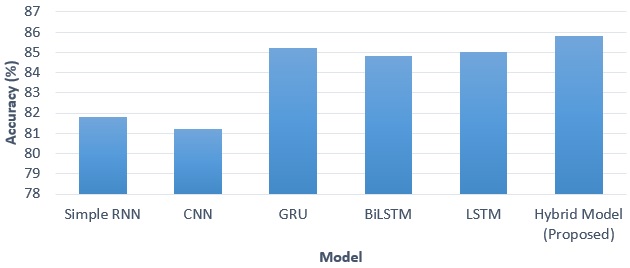}
\caption{Performance comparison of different models.}
\label{accuracy_comparison}
\end{figure}
\begin{table}[]
\centering
\caption{Evaluation of metrics of different models.}
\label{metrics}
\begin{tabular}{|c|c|c|c|}
\hline
\textbf{Model}                                                                & \textbf{Precision} & \textbf{Recall} & \textbf{F1-score} \\ \hline
Simple RNN                                                                    & 0.82               & 0.82            & 0.82              \\ \hline
CNN                                                                           & 0.82               & 0.81            & 0.81              \\ \hline
GRU                                                                           & 0.85               & 0.85            & 0.85              \\ \hline
BiLSTM                                                                        & 0.85               & 0.85            & 0.85              \\ \hline
LSTM                                                                          & 0.85               & 0.85            & 0.85              \\ \hline
\textbf{\begin{tabular}[c]{@{}c@{}}Hybrid Model\\    (Proposed)\end{tabular}} & 0.86               & 0.86            & 0.86              \\ \hline
\end{tabular}
\end{table}
We compare the sentiment analysis method in the proposed framework with the basic deep learning models such as SimpleRNN, CNN, GRU, BiLSTM, and LSTM using standard parameters in the dataset~\cite{sarker2021deep}. By implementing the word2vec (skip-gram version), we also train different deep learning models for the dataset and compare their results with the proposed hybrid model in the framework. Fig~\ref{accuracy_comparison} represents the results of accuracy comparisons and Table~\ref{metrics} shows the precision, recall, and f1-score comparisons of different models. From Fig.~\ref {accuracy_comparison}, it can be seen that almost all deep deep learning models provide good results with an accuracy level of more than 80\%. The values reveal that our proposed approach outperforms other models and provides about 86\% of accuracy. From the results of Table~\ref{metrics}, it can be seen that the performance of the proposed hybrid model is higher compared to other basic deep learning models in terms of precision, recall, and f1-score for the weighted average of negative, neutral, and positive sentiment predictions. The experimental results show that the classification performance of the proposed hybrid deep learning classifier is better with the input of aspect terms and sentiment terms in the GRU branch of the model. 
\subsubsection{Error Analysis}
\begin{figure}
\includegraphics[width=22.5pc,height=5cm]{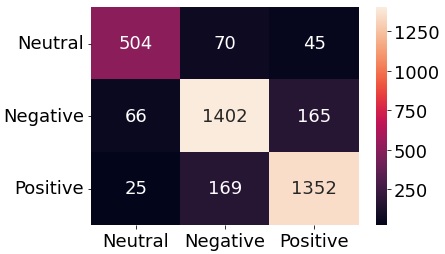}
\caption{Confusion matrix of proposed hybrid deep learning model.}
\label{confusion_matrix}
\end{figure}
The confusion matrix is used to perform the detailed error analysis as shown in Fig.~\ref{confusion_matrix}. It is clear from the matrix that few data are incorrectly classified. For example, 45 out of 619 cases of the neutral class were predicted positive. In the negative class, 66 of the 1633 data were mistakenly classified as neutral. Error analysis shows that the positive class achieved the highest rate of accurate classification (87.45 \%) while the neutral class achieved the lowest (81.42 \%). A possible reason for the high incorrect prediction in the neutral class may be the presence of a small number of neutral class tweets in the training dataset. However, sentiment analysis is widely subjective, depends on individual perceptions, and people may consider a tweet in many ways. Therefore, by creating a balanced dataset with a variety of high amounts of data, the incorrect predictions can be reduced to some degree.
\section{Discussion}
\begin{figure}

\centering
\includegraphics[width=12cm,height=8cm]{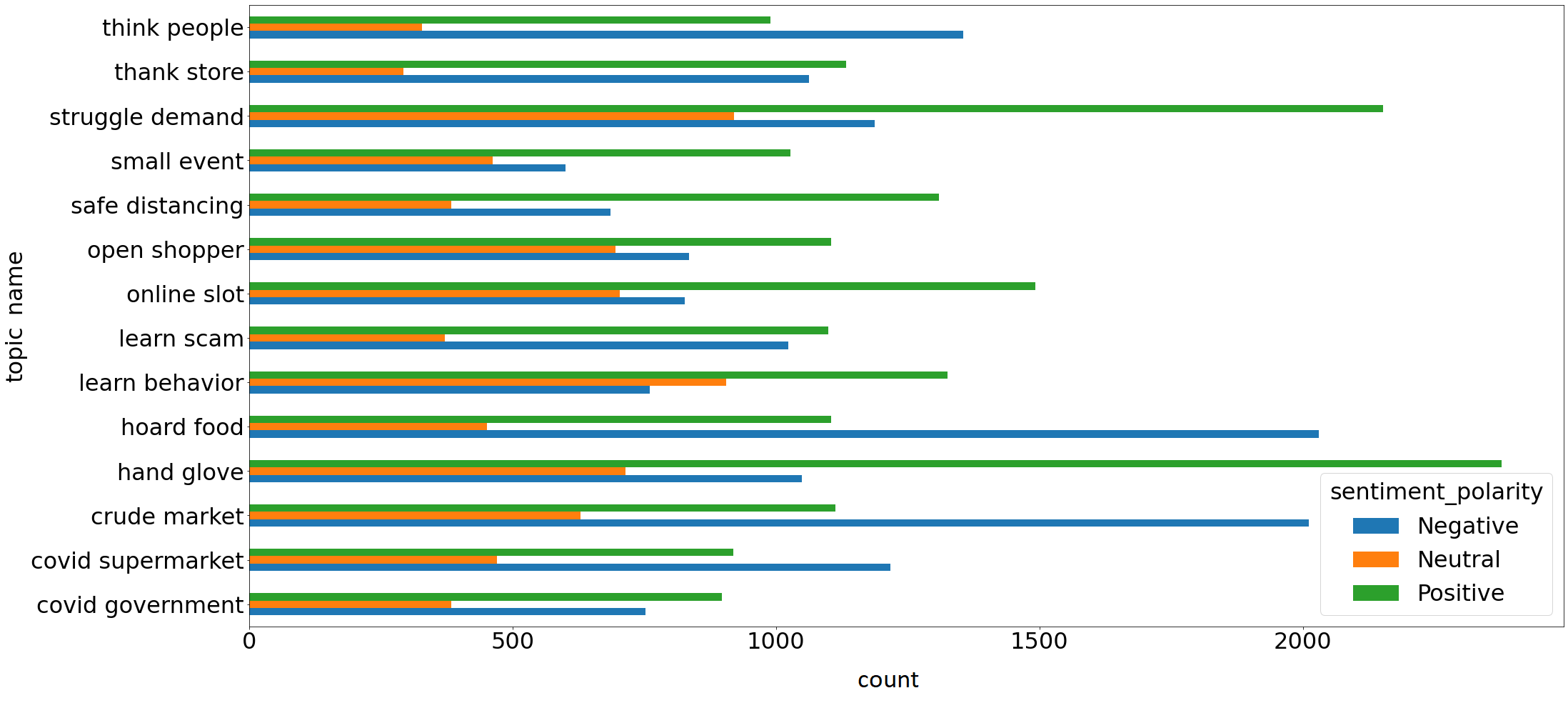}
\caption{Topic based sentiment analysis.}
\label{topic_based_sentiment}

\end{figure}
We present topic-based sentiment analysis to extract useful information from COVID-19-related tweets. We plot and count the number of positive, neutral, and negative tweets for each of the fourteen topics as shown in Fig.~\ref{topic_based_sentiment}. The chart reports that in each topic most of the tweets are positive. In topics like covid supermarket, crude market, hoard food and think people we find that number of tweets that carry negative sentiment is higher. The topic-based sentiment analysis related to the COVID-19 tweets helps to understand people's perceptions and emotions for taking necessary steps to reduce suffering and compensate for the loss of resources. Government and policymakers can therefore take the necessary steps to understand the details of the human problems and their needs on the basis of public sentiments that reflect tweets on social media in order to minimize the detrimental effects of the COVID-19 pandemic.

Overall, in this paper, we present a data-driven~\cite{sarker2021data} framework for the topic-based sentiment analysis to extract beneficial information from the COVID-19 related Twitter dataset. We use a laptop with a 2.0 GHz Core i3 processor and 4 GB RAM to execute our proposed technique. We write codes in the Python programming language and run our codes within the Google Colaboratory platform. We firmly believe that the proposed framework can be used effectively in other areas of applications such as agriculture, health, education, business, cyber security, etc. It requires a very large dataset to train a transformer from scratch like BERT, ALBERT, and RoBERTA to achieve better results. Hence, in this paper, we use classical deep learning classifiers such as BiLSTM and GRU to build the hybrid model~\cite{ezen2020comparison}. In the future, we will collect more data and expand our scope of experiments through the use of above mentioned models in the field of topic-based sentiment analysis. We will also apply our proposed framework in other languages and use it on other social media platforms at various events to generate valuable information to dynamically handle the ever-increasing data load.
\section{Conclusion}
In this paper, we have presented a new topic-based sentiment analysis framework that effectively and automatically identifies key topic labels and corresponding sentiments from COVID-19 tweets. We have used the popular topic modeling algorithm LDA to extract hidden topics from tweets. Then we have used unigrams of the sentiment terms and aspect terms of the tweets to produce significant and meaningful topic labels on the basis of soft cosine similarity (SCS) values. Our proposed topic labeling method performs better and helps to categorize a huge amount of tweets corresponding to semantically similar topic labels to highlight user conversations.

In this paper, we have also presented a hybrid deep learning model for sentiment analysis. The hybrid model proposed in the framework uses the benefits of GRU, BiLSTM, and GAP models as explained in the methodology section of this paper. Our proposed hybrid deep learning model in the framework achieves better results compared to other benchmark models. We have also categorized sentiment polarities towards each topic. Our proposed framework effectively facilitates topic-based sentiment analysis based on COVID-19 tweets and detects various issues related to the COVID-19 pandemic. Our framework saves time and reduces the human effort to reduce the overhead of complex topic labeling and sentiment analysis activities to get a broader public opinion on social media platforms such as Twitter. We believe that the proposed framework will help policymakers to identify the various problems associated with COVID-19 by analyzing tweets.

\section*{Declarations}
\textbf{Conflict of Interests:} The authors declare no conflict of interest. \\
\textbf{Ethical statements:}
The authors follow all the relevant ethical rules. \\
\textbf{Data availability statement:} Data used in this work can be made available upon reasonable request. \\
\textbf{Funding:} Not Applicable.

\nolinenumbers


%
%
%



\end{document}